\documentclass[runningheads]{llncs}
\usepackage{graphicx}

\usepackage{tikz}
\usepackage{comment}
\usepackage{amsmath,amssymb} 
\usepackage{color}

\def\degree{${}^{\circ}$}
\usepackage{multirow}

\begin{document}
\pagestyle{headings}
\mainmatter
\def\ECCVSubNumber{5287}  

\title{Object Level Depth Reconstruction for Category Level 6D Object Pose Estimation From Monocular RGB Image} 



\titlerunning{Abbreviated paper title}

\author{Zhaoxin Fan\inst{1}\and
Zhenbo Song\inst{2} \and
Jian Xu\inst{4}\and
Zhicheng Wang\inst{4}\and
Kejian Wu\inst{4} \and
Hongyan Liu\inst{3} \and
Jun He\inst{1}  \thanks{Corresponding author} 
}

\authorrunning{Zhaoxin et al.}
%
\institute{Key Laboratory of Data Engineering and Knowledge Engineering of MOE,
School of Information, Renmin University of China, 100872, Beijing, China\\
\email{\{fanzhaoxin,hejun\}@ruc.edu.cn}\and
School of Computer Science and Engineering, Nanjing University of Science and Technology, 210094, Nanjing, China\\
\email{songzb@njust.edu.cn}\and
Department of Management Science and Engineering, Tsinghua University, 100084, Beijing, China\\
\email{hyliu@tsinghua.edu.cn}\and
Nreal\\
\email{\{jianxu,zcwang,kejian\}@nreal.ai}}

\maketitle
\begin{sloppypar}
\begin{abstract}
Recently, RGBD-based category-level 6D object pose estimation has achieved promising improvement in performance, however, the requirement of depth information prohibits broader applications. In order to relieve this problem, this paper proposes a novel approach named \textbf{O}bject \textbf{L}evel \textbf{D}epth reconstruction \textbf{N}etwork (\textbf{OLD-Net}) taking only RGB images as input for category-level 6D object pose estimation. We propose to directly predict object-level depth from a monocular RGB image by deforming the category-level shape prior into object-level depth and the canonical NOCS representation. Two novel modules named Normalized Global Position Hints~(NGPH) and Shape-aware Decoupled Depth Reconstruction~(SDDR) module are introduced to learn high fidelity object-level depth and delicate shape representations. At last, the 6D object pose is solved by aligning the predicted canonical representation with the back-projected object-level depth. Extensive experiments on the challenging CAMERA25 and REAL275 datasets indicate that our model, though simple, achieves state-of-the-art performance.


\end{abstract} 
\begin{figure}[t]
 \centering
  \includegraphics[width=0.5\linewidth]{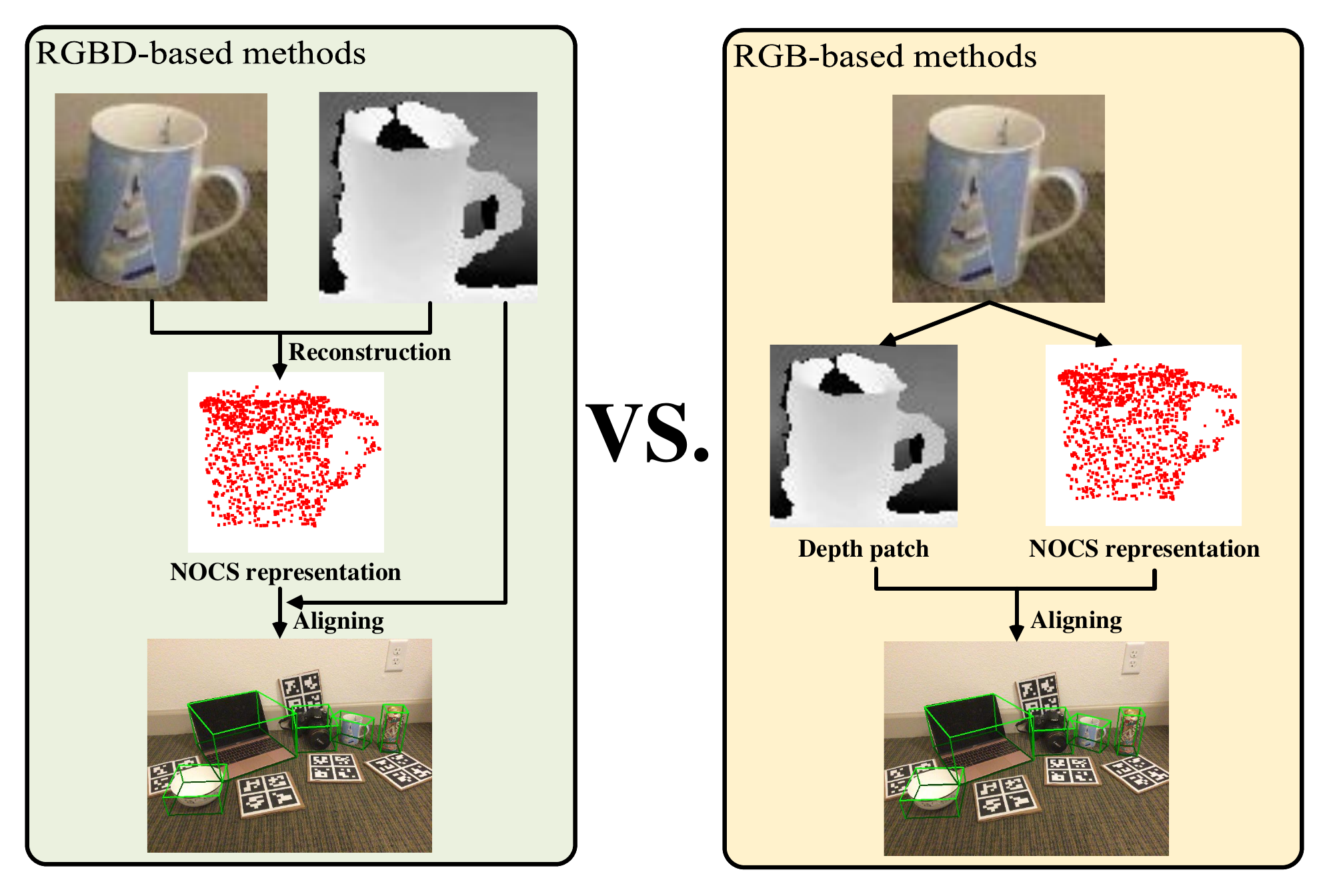}

  \caption{Difference between RGBD-based methods and our RGB-based method. RGBD-based methods take RGB image and depth channel as inputs, and the output is a canonical NOCS representation. While our RGB-based method only takes RGB images as input, and predict the NOCS representation as well as the object-level depth simultaneously. }

  \label{fig:idea}
  
\end{figure}
\section{Introduction}
\label{sec:intro}
Category-level 6D object pose estimation that predicts the full degrees of rotation and translation of an object w.r.t the camera is an significant yet challenging task. It has a wide range of applications including robotic grasping \cite{tremblay2018deep,du2019vision,wang2020feature}, augmented reality \cite{su2019deep,tan20176d,rambach20186dof,zhao2022tracking}, and autonomous driving \cite{sun2020scalability,caesar2020nuscenes,grigorescu2020survey,xu2022v2x,xu2021opv2v}.  As shown in Fig. \ref{fig:idea} left, the mainstream framework of this task includes two steps :  1) Predicting a canonical representation to represent the canonical category-level object shape. 2) Aligning the canonical representation with the back-projected object depth to predict the object pose. Wang et al. \cite{wang2019normalized} first introduces a model named NOCS to do this task. Then, many following works \cite{wang2019normalized,chen2020learning,tian2020shape,chen2021fs,lin2021dualposenet} are proposed to improve the performance from various perspectives, such as, canonical representations \cite{chen2020learning}, learning shape priors \cite{tian2020shape}, and new augmentation strategies \cite{chen2021fs}. All these methods have achieved promising object pose estimation accuracy. Nevertheless, they are all RGBD-based, whose performance are dominated by features learned from the depth channel. Since the depth information is always not available in many scenarios, it seriously limits broader applications for these methods. Lee et al.~\cite{lee2021category} proposes to take only  monocular RGB image as input for category-level 6D object pose estimation. It predicts a depth map from the RGB image for information reinforcement. One limitation of Lee et al.~\cite{lee2021category} is that it needs to reconstruct a metric-scale mesh for each object first, making the pipeline redundant. In this paper, we explore to directly predict the object-level depth from an input RGB image patch without relying on any intermediate representation, for instance, a mesh, in a simple yet effective way.


This paper proposes a novel approach named \textbf{O}bject \textbf{L}evel \textbf{D}epth reconstruction \textbf{N}etwork (\textbf{OLD-Net}) for RGB-based category-level 6D object pose estimation. Fig. \ref{fig:idea} right illustrates the main pipeline of the OLD-Net. In detail, object-level depth and the NOCS representation are simultaneously predicted from the input RGB image, and aligned together to predict the 6D object pose. Different from the previous method \cite{lee2021category} that predicts the depth of the object region by reconstructing a mesh, the object's observed depth is directly predicted from the RGB image in an end-to-end manner in this paper.  

To obtain the depth of the object region, a straight-forward way is to predict the scene-level depth. However, due to variety and diversity of field of view, the predicted scene-level depth is usually coarse, resulting in the loss of object shape details. The pose estimation performance would also suffer from it (as shown in Table \ref{comapre:scene}). To release the problem, we propose to reconstruct the object-level depth directly by learning to deform category-level shape priors. In contrast to predict a scene-level depth map, reconstructing the object-level depth is more computational friendly and can preserve better shape details, benefiting the subsequent depth-NOCS alignment process.

To delicately reconstruct the object-level depth, a novel module named Normalized Global Position Hints~(NGPH) is proposed in OLD-Net to balance scene-level global information and local feature-level shape details. NGPH is the normalized 2D detection results with camera intrinsics, providing global position cues about the object's absolute depth in the scene, as well as the generalization ability towards images captured by different cameras. Furthermore, a Shape-aware Decoupled Depth Reconstruction (SDDR) scheme is utilized to predict the shape details and absolute depth. In particular, SDDR decouples and predicts the absolute depth into shape points and depth translation with two independent deep networks. Intuitively, the shape points are to preserve shape details while the depth translation is designed for predicting absolute object center. 


Apart from depth, we further predict the NOCS representation \cite{wang2019normalized} of the target object following the RGBD-based methods \cite{wang2019normalized,tian2020shape}. A discriminator is utilized during training to improve reconstruction quality. We back-project the object-level depth into a point cloud after both the NOCS representation and the observed object-level depth are predicted. They are aligned by a Umeyama algorithm \cite{umeyama1991least} to solve for the 6D object pose as shown in Fig. \ref{fig:idea}. We conduct extensive experiments on the famous CAMERA25 \cite{wang2019normalized} dataset and REAL275 \cite{wang2019normalized} dataset. Experimental results demonstrate that our method achieves state-of-the-art performance.

Our main contributions are:
\begin{itemize}
\item We propose OLD-Net, a novel deep learning approach for category-level 6D object pose estimation, which aims at directly predicting object-level depth from a monocular RGB image in a simple yet effective way.

\item  We propose the Normalized Global Position Hints and the Shape-aware Decoupled Depth Reconstruction scheme. Both modules are tailored for RGB-based category-level 6D object pose estimation.

\item We conduct extensive experiments on two challenging datasets to verify the effectiveness of our method. Our model achieves state-of-the-art performance in both synthetic and real world scenarios.
\end{itemize}

\section{Related Work}

\subsection{Instance-level 6D Object Pose Estimation}
 We broadly classify instance-level 6D object pose estimation methods into RGB-based methods and RGBD-based methods according to input data format. For RGB-based methods, PoseNet \cite{kendall2015posenet} is a pioneering network that introduces a CNN architecture to directly regress a 6D camera pose from a single RGB image. Then, PoseCNN \cite{xiang2017posecnn} proposes to first locate the object in the image then predicts the depth of the object for more accurate pose estimation. However, directly estimating the object pose from an RGB image is challenging due to the non-linearity of the rotation space. Therefore, to overcome this, methods like \cite{tekin2018real} propose to first predict 2D keypoints as intermediate representations and then solve a PnP problem \cite{lepetit2009epnp} to recover the accurate object pose. To improve keypoint detection performance, PVNet \cite{peng2019pvnet} formulates a voting scheme, which is more robust towards occlusion and truncation. Hybridpose \cite{song2020hybridpose} improves PVNet by adding more constraints such as edges. In contrast to predicting sparse keypoints, methods like DPOD \cite{zakharov2019dpod} predict dense-correspondence for 6D object pose estimation. There are also some methods like \cite{li2018deepim} propose to refine the predicted object pose through iteratively comparing observed and rendered images. For RGBD-based methods, PVN3D \cite{he2020pvn3d} proposes a Hough voting network to predict 3D keypoints. The object pose is recovered by aligning these keypoints with the object models using ICP. Wang et al. \cite{wang2019densefusion} introduce DenseFusion, which densely fuses RGB features and point cloud features for 6D object pose detection. MoreFusion \cite{wada2020morefusion} uses a more compact fusion strategy for further performance improvement.

Though having achieved promising performance, instance-level methods suffer from limited generalization ability because one model only works for a particular object in their settings. In this paper, we study the more general category-level 6D object pose estimation.

\subsection{Category-level 6D Object Pose Estimation}

NOCS \cite{wang2019normalized} may be the first work that uses deep learning models for category-level 6D object pose estimation. The prevalent CAMERA25 and REAL275 datasets are released in NOCS. The canonical NOCS representation are used to recover the 6D object pose. Chen et al. \cite{chen2020learning} propose another canonical representation named CASS and this work also designs a deep learning approach to cope with this task. SPD \cite{tian2020shape} claims that the NOCS representation's potential is not fully explored yet. Therefore, it presents the Shape Prior Deformation idea to better predict the NOCS representation. We also use the Shape Prior Deformation, however, to predict the object-level depth, apart from predicting the NOCS representation. Chen et al. \cite{chen2021fs} propose a novel augmentation method in FS-Net to improve training quality. Lin et al. \cite{lin2021dualposenet} introduce a dual network architecture in their work DualPoseNet to improve the 6D object pose estimation performance. One observation is that most of category-level methods solving the pose estimation problem by a estimation-by-aligning scheme, which may partly due to that alternative solutions like PnP+RANSAC or direct pose parameters prediction are hard to deal with ambiguities caused by shape variations. In this paper, we also follow the mainstream of estimation-by-aligning scheme.

Though with good performance, the above-mentioned methods all need to take an RGBD observation as input, which severely limits their application scenarios. To the best of our knowledge, \cite{lee2021category} is the only work tailored for RGB-based category-level 6D object pose estimation. It first uses Mesh-RCNN \cite{gkioxari2019mesh} to predict a metric-scale object mesh. Then the mesh is rendered into a depth map for pose estimation. Our method is significantly different from \cite{gkioxari2019mesh}. First, we propose a more simple way to directly predict the object-level depth from the monocular RGB image. Second, we design the novel SDDR and NGPH. Third, we verify the importance of using shape prior to better cope with intra-class variation. The object-level depth predicted by our approach is of higher quality, significantly improves the 6D pose estimation performance.

\begin{figure*}[t]
 \centering
  \includegraphics[width=0.8\linewidth]{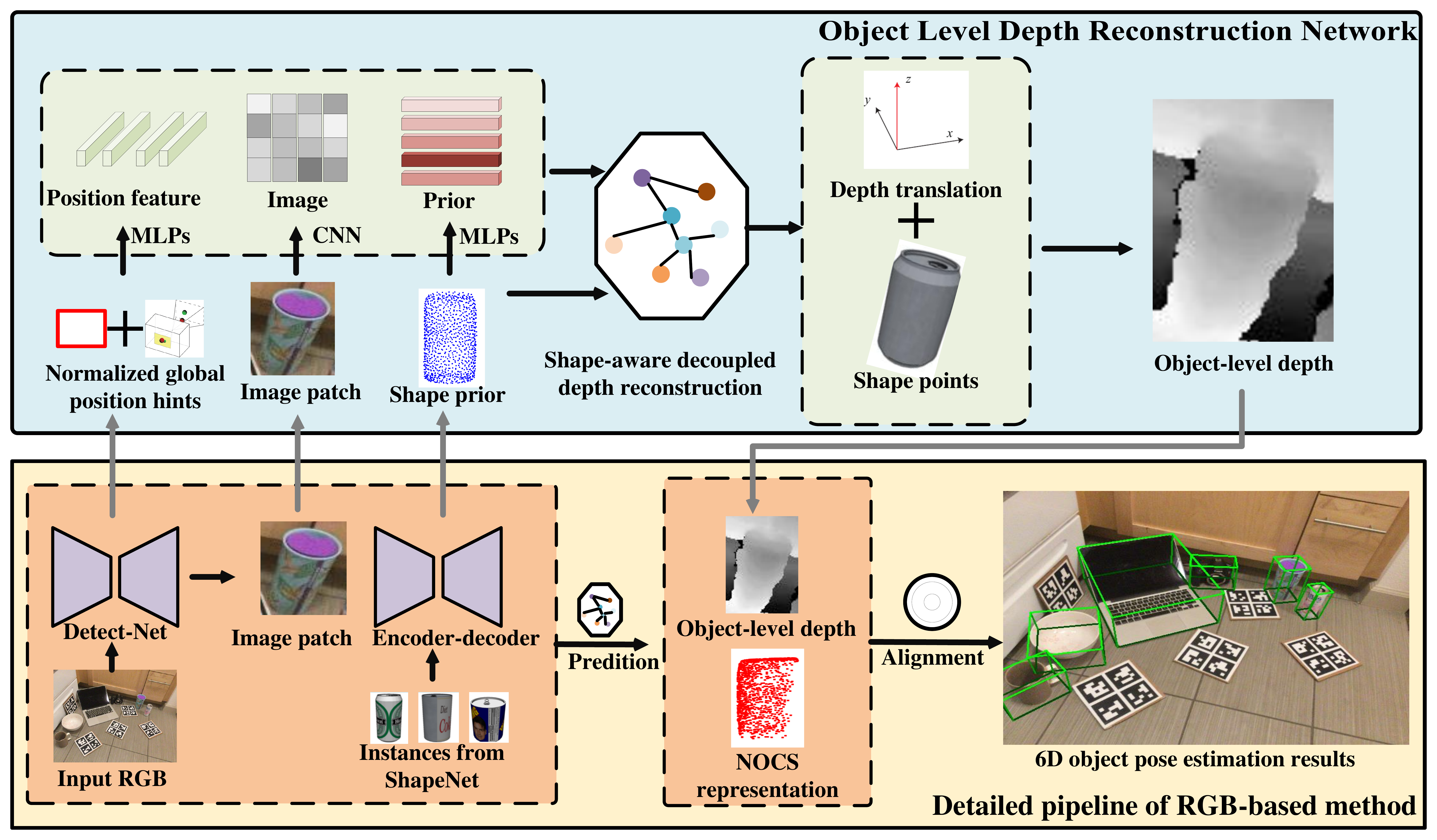}

  \caption{ The bottom figure is the pipeline of this paper, we first train a Detect-Net and an encoder-decoder network to crop image patches and generate the shape prior respectively. Then, we predict the object-level depth and the NOCS representation. 6D object pose is recovered by aligning the depth and the NOCS representation. To predict high-quality object-level depth, we design the novel OLD-Net as shown in the top figure. }
  \label{fig:pipeline}

\end{figure*}
\section{Pipeline}
 We illustrate the pipeline of our work in Fig. \ref{fig:pipeline} (bottom). The network architecture of our main contribution OLD-Net is shown in Fig. \ref{fig:pipeline} (top).

Our pipeline takes an image patch $I_{patch} \in R^{W \times H \times 3}$ and a shape prior $P_{pri} \in R^{N_m \times 3}$ as input.  The image patch is cropped by a trained detector called Detect-Net, which represents object specific information. The shape prior of the target category is predicted by an encoder-decoder network using the mean embedding, which is used to alleviate category-variation \cite{tian2020shape}.


 Subsequently, the image patch and shape prior are input into the proposed OLD-Net  to reconstruct the object-level depth as shown in Fig. \ref{fig:pipeline} top. Besides, OLD-Net also takes the 2D detection results from the Detect-Net as well as the camera intrinsics as inputs, which are normalized as NGPH.  A Shape-aware Decouple Depth Reconstruction scheme is occupied in OLD-Net to preserve shape details and absolute object center.

Finally, The NOCS representation \cite{wang2019normalized} is predicted using a deep network following \cite{tian2020shape}. Then, we back-project the object-level depth into a point cloud and adopt the Umeyama algorithm \cite{umeyama1991least} to recover the object pose as in \cite{wang2019normalized}.

Next, we first introduce the OLD-Net at length. Then, we describe how we predict the NOCS representation. Finally, we introduce the loss function used in this paper.

\section{OLD-Net}
The most serious difficulty for category-level object pose estimation is how to deal with shape variations. RGBD-based methods like \cite{tian2020shape} have demonstrated that directly predicting the canonical NOCS representation is not promising. Therefore, for each category, they learn a category-level shape prior as an intermediate representation to ease the task. Compared to directly estimating the NOCS representation, problems caused by shape variation can be largely mitigated by utilizing the learned shape prior. In our work, we follow this framework to build OLD-Net owing to shape-prior's superiority. Further more, we extend it to predict the object-level depth. In addition, two novel modules technically designed for RGB-based methods named NGPH and SDDR are proposed to better capture useful information.

In particular, as shown in Fig. \ref{fig:pipeline} top, taking the image patch, the shape prior, and the NGPH as input, OLD-Net first uses two MLPs and a CNN to learn high-level image feature $I_p \in R^{W \times H \times C}$, prior feature $f_{pri} \in R^{ N_m \times C}$, and position feature $f_{pos} \in R^{2C}$ first. Then, utilizing these features, the shape points as well as the depth translation are concurrently predicted by the Shape-aware Decoupled Depth Reconstruction (SDDR) scheme. Lastly, the shape points and depth translation are reassembled together to obtain the object-level depth. Next, we introduce NGPH and SDDR in detail.

\subsection{Normalized Global Position Hints}

\label{sec:postioncues}
We are motivated that the object-level depth can be directly predicted from an image with high quality. The most straightforward way to achieve this goal is to predict a scene-level depth map. However, predicting the scene-level depth map from the raw image is computational costly. Furthermore, it may cause the loss of object's shape detail. Note the shape detail is very important for our pipeline since we need to align two 3D representations to recover the object pose. Therefore, we choose to take the image patch of a particular object as input to predict the object-level depth. Nevertheless, the image patch has lost the absolute global position information of the object due to the crop and resize operation, resulting in the predict depth being suffered from scale ambiguities. To this end, we propose the NGPH, which properly solves this problem by providing absolute global position information and solving scale ambiguities.

Inspired by \cite{song2020end}, we choose the parameters of the 2D bounding box $(l,t,r,b)$ output by the Detect-Net to form NGPH, which represents the left, top, right, and bottom image coordinates of the 2D bounding box. This information is effective enough to be supplied to the network to infer scale clues, i.e. recovering the absolute object depth if all images are captured by the same camera. However, it is common that images would be collected by different cameras. And it is a common sense that the absolute object depth would be influenced by the camera intrinsic if it is inferred from a single monocular RGB image. Therefore, we propose to inject the camera intrinsics into NGPH. Hence, the trained network can also be generalized into other images captured by different cameras. We make use of the 2D bounding box and the camera intrinsics by normalizing them into canonical coordinates:
\begin{equation}
g=[\frac{f_x}{r-l},\frac{f_y}{b-t},\frac{l-c_x}{f_x},\frac{t-c_y}{f_y},\frac{r-c_x}{f_x},\frac{b-c_y}{f_y}]
\end{equation}
where $g$ represents the final NGPH we use, $c_x$ and $c_y$ represent coordinates of the camera optical center, and $f_x$ and $f_y$ represent focal lengths. The first two terms normalize the size of the object's bounding box with the focal length, eliminating scale ambiguity caused by object size. The later four terms normalize the center of the object using the focal length and size of the bounding box, removing ambiguity but preserving the position information. 


The proposed NGPH, validated by experimental results that, though simple, is indispensable in object-level depth reconstruction.

\subsection{Shape-aware Decoupled Depth Reconstruction}
\label{sec:depth}
We apply an estimation-by-aligning scheme to recover the 6D object pose in our pipeline. Therefore, we have to estimate the observed 3D representation of the object first. In our work, we choose to reconstruct the object-level depth to form the "observed" 3D representation. We hope the reconstructed object-level depth to be as similar as the one captured by a depth camera as possible. Besides, the object's shape detail should be well described because we need to align the depth with the NOCS representation for recovering accurate 6D object poses. To achieve so, we propose a SDDR scheme. Specifically, in SDDR, we propose to decouple the object-level depth into shape points and depth translation. The former preserves relative position information between "observed" points, where the shape detail of the observed object is expected to be delicately predicted along with it. While the latter describes the positional information of the object center. We use two different modules to predict them separately. 

The features used in the OLD-Net are the reshaped image feature matrix $f_I \in R^{N_p \times C}$( where $N_p$ is the number of pixels), the position feature $f_{pos} \in R^{2C}$ and the prior feature $f_{pri} \in R^{ N_m \times C}$. We also apply MLPs and adaptive average pooling to obtain the global image feature $f^g_{I} \in R^{C_g} $ and the global prior feature $f^g_{pri} \in R^{C_g}$.
 
\noindent \textbf{Shape points prediction} We adopt the Shape Prior Deformation (SPD) idea \cite{tian2020shape,fan2021acr} to reconstruct shape points, which would provide the model with more constraints on object shape. Specifically, utilizing the above features, the network would learn a deform field $D_{depth} \in R^{N_m \times 3}$ and an assign field $M_{depth} \in R^{N_n \times N_m}$ to deform and assign the shape prior into the back-projected point cloud of the object-level depth:
\begin{equation}
P_{depth}=M_{depth}(P_{pri}+ D_{depth})
\end{equation}
To learn $D_{depth}$, we repeat $f^g_{I}$, $f^g_{pri}$ and $f_{pos}$ for $N_m$ times and concatenate them with $f_{pri}$. The concatenated features are input into a MLP to learn $D_{depth} $. Similarly, to learn $M_{depth}$, we repeat $f^g_{I}$, $f^g_{pri}$ and $f_{pos}$ for $N_p$ times and concatenate them with $f_I$. Another MLP is used to learn $M_{depth}$.  Note that in the literature of category-level tasks \cite{fan2021deep}, shape prior is easily accessible and is widely employed for predicting canonical representations. In this paper, we use it to predict object-level depth to provide a guidance for future RGB-based works. 

\noindent \textbf{Depth translation prediction} To learn the absolute position of the object center, we propose to use an independent head to learn the depth translation, which is implemented as a MLP. We use the concatenation of $f_{pos}$ and $f^g_{I}$ as the input. The output is a single value that represents the absolute depth of the object center. We name it as $Z_t$. 

The SDDR scheme mainly benefits from three aspects to preserve object shape details. First, since we only take image patches to reconstruct the object-level depth, the model can concentrate on the object's shape rather than the whole scene geometry. Second, the shape prior provides powerful constraints about the shape of the object, making it easier to recover shape details. Third, absolute object center and object shape are learned separately and would be paid with different attention.

After both $P_{depth}$ and $Z_t$ are predicted. The object-level depth can be represented as $Z=Z_{depth}+Z_{t}$, where $Z_{depth}$ is the third component of $P_{depth}$. Note that we choose to supervise $Z$ rather than the back-projected point cloud. It is because on the one hand, it is easier for the network to learn $Z$ during training, on the other hand, back-projecting $Z$ to a point cloud for aligning, the 2D coordinates of the object would provide additional constraints for the global position, which would also benefit the final pose recovery step.

\section{NOCS Representation Prediction}
\label{sec:nocs}

We also predict the NOCS representation \cite{wang2019normalized} of the target object in our pipeline, which is a canonical representation that is used to align with the object-level depth to recover the 6D object pose. To predict the NOCS representation, we back-project $Z$ into a point cloud and input it into a MLP to learn depth features $f_{depth} \in R^{N_p \times C}$. Taking $f_{depth}$, $f_{pri}$ and $f_I$ as input, similar to reconstruct the object-level depth, we use SPD to predict the NOCS representation:

\begin{equation}
P_{nocs}=M_{nocs}(P_{pri}+ D_{nocs})
\end{equation}

However, we find that in some cases, $P_{nocs}$ is not realistic enough, which would affect the final 6D object pose estimation accuracy. Therefore, we adopt the adversarial training strategy \cite{ledig2017photo} to train the network. Specifically, we design a discriminator $\mathbb{D}$ to judge whether the predicted NOCS representation is real enough or not. The optimization goal of the discriminator can be represented as:
\begin{equation}
L_d=(\mathbb{D}(\hat{P_{nocs}})-1)^2+(\mathbb{D}({P_{nocs}}))^2
\end{equation}

where $\hat{P_{nocs}}$ is the ground-truth NOCS representation.

Similarly, the optimization goal of the NOCS prediction network is $P_{nocs}$ is: 
\begin{equation}
L_g=(\mathbb{D}({P_{nocs}})-1)^2
\end{equation}

During training, we iteratively update the parameters of the discriminator and the NOCS prediction network. Both networks would become stronger and stronger through confrontation. Therefore, the predicted NOCS representation would also become more and more realistic.

\section{Loss Function}
\label{sec:loss}
For the object-level depth reconstruction, we use the $L1$ loss between $Z$ and the ground-truth:
\begin{equation}
L_z=|z-\hat{z}|_1
\end{equation}

For the NOCS representation prediction, we use the same loss function as \cite{tian2020shape} due to its excellent performance. This loss function includes a smooth L1 loss $L_{corr}$ between the reconstructed NOCS representation and ground-truth to encourage better one-to-one correspondence, the chamfer distance loss $L_{cd}$ between the deformed shape prior $P_{pri}+D_{nocs}$ and the object's canonical instance model to preserve object shape, a cross-entropy loss $L_{entro}$ to encourage peak distribution of the assignment matrix $M_{nocs}$,  and a L2 regularization loss $L_{reg}$ on $M_{nocs}$ to avoid collapsing deformation.

The overall loss function we use is:
\begin{equation}
L=\gamma_1L_z+\gamma_2L_d+\gamma_3L_g+\gamma_4L_{corr}+\gamma_6L_{cd}+\gamma_6L_{entro}+\gamma_7L_{reg}
\end{equation}

where $\gamma_1$ to $\gamma_7$ are balance terms.

\section{Experiment}
To verify the effectiveness of our method. We conduct experiments on the CAMERA25 dataset \cite{wang2019normalized} and REAL275 dataset \cite{wang2019normalized}. More details about implementation, datasets and metrics can be found in the SuppMat.

\begin{table*}[t]
		\centering
		\begin{tabular}{c |c|c|c|c|c|c|c } 
			\hline 	
			Data& Reconstructions & bottle& bowl& camera& can& laptop& mug\\
	  \hline 
		\multirow{2}*{CAMERA25}&Depth&0.0532& 0.0360&0.0596 &0.0407 &0.0371&0.0422 \\
			& NOCS &	0.0225	&0.0142	&0.0192&	0.0251	&0.0150	&0.0178\\
		\hline
		\hline
		\multirow{2}*{REAL275}& Depth &0.0179 &0.0200& 
0.0136&0.0162&0.0151& 0.0129 \\
			& NOCS &	0.0247	&0.0174	&0.0215&	0.0205	&0.0203	&0.0203\\
   \hline 

	\end{tabular}
	\caption{Reconstruction quality evaluation. The evaluation metric is chamfer distance.}
	\label{reconstructionquality}

\end{table*}

\begin{table*}[t]
		\centering
		\begin{tabular}{c |c|c|c|c|c|c|c } 
			\hline 	
			Data& Methods & IoU25& IoU50& IoU75& 10 cm & 10\degree& 10\degree 10cm\\
	  \hline

					\multirow{2}*{CAMERA25}& lee et al.\cite{lee2021category}&\textbf{75.5}&\textbf{32.4} &5.1 &29.7 &60.8& 19.2 \\
			& OLD-Net(Ours) &74.3&	32.1	&\textbf{5.4}&\textbf{30.1}	&\textbf{74.0}&\textbf{23.4}\\
						\hline
			\hline
				\multirow{3}*{REAL275}& Synthesis \cite{chen2020category}&-&-
 &-&34.0 &14.2& 4.8 \\
		
		& lee et al.\cite{lee2021category}&62.0&23.4
 &\textbf{3.0} &\textbf{39.5} &29.2& 9.6 \\
			& OLD-Net(Ours) &\textbf{68.7}&	\textbf{25.4}	&1.9&38.9	&\textbf{37.0}&\textbf{9.8}\\
   \hline 

	\end{tabular}
	\caption{ Quantitative comparison with state-of-the-art methods on the CAMERA25 and REAL275 dataset.}
	\label{comapre:real}

\end{table*}

\begin{table*}[t]
		\centering
		\begin{tabular}{c |c|c|c|c|c|c|c } 
			\hline 	
			Data& Methods & IoU25& IoU50& IoU75& 10 cm & 10\degree& 10\degree 10cm\\
	  \hline

\multirow{3}*{CAMERA25}& Scene-level baseline (shared)&47.7&12.9 &0.8 &17.8 &8.0& 1.6 \\
&Scene-level baseline (independent) &50.4&14.0 &0.9 &17.7 &10.0& 2.2 \\
			& OLD-Net(Ours) &\textbf{74.3}&	\textbf{32.1}	&\textbf{5.4}&\textbf{30.1}	&\textbf{74.0}&\textbf{23.4}\\
						\hline

   \hline 

	\end{tabular}
	\caption{ Quantitative comparison with state-of-the-art scene-level depth prediction baselines on the CAMERA25 dataset.}
	\label{comapre:scene}

\end{table*}

\subsection{Results}
\textbf{Reconstruction quality evaluation}:
In our work, the main idea is to reconstruct the object-level depth and the NOCS representation. Therefore, we first evaluate the reconstruction quality of our method in Table \ref{reconstructionquality}. We compute the chamfer distance between back-projected depth and the ground-truth to validate the depth reconstruction quality. We also compute the chamfer distance between the predicted and ground-truth NOCS representations to evaluate the NOCS prediction quality. It can be seen from Table \ref{reconstructionquality}, for object-level depth reconstruction, the errors are less than 2cm in the REAL275 dataset. For the NOCS prediction, the error is also close to 2cm in the REAL275 dataset. 2cm is a relatively small scale error compared to our large object size and scene depth. Therefore, we can conclude that our method indeed achieves good object-level depth reconstruction quality and NOCS representation prediction quality. On the CAMERA25 dataset, the NOCS representation prediction error is still below 2cm for most of the categories. However, the object-level depth construction error is increased to 3cm to 5cm. The reason may be that there exists a larger depth distribution variance in the larger synthetic dataset. This observation also indicates that reconstructing object-level depth is harder than predicting the NOCS representation.

\noindent \textbf{Quantitative results of 6D pose estimation}: We quantitatively compare our method with state-of-the-art methods in Table \ref{comapre:real}. We first compare our method with Lee et al. \cite{lee2021category} on the CAMERA25 dataset. Lee et al. \cite{lee2021category} predict the depth by first reconstructing a mesh and then rendering the mesh into a depth map. In contrast, we choose to directly reconstruct the object-level depth, which is more simple yet more effective. We can see that our method outperforms Lee et al. \cite{lee2021category} at 4 metrics among the 6. In the most strict 10 \degree 10cm metric, our method exceeds Lee et al. \cite{lee2021category} for 4.2 points, which is a significant improvement. On the IoU25 and IoU50 metrics, though our results are slightly lower than Lee et al. \cite{lee2021category}, we still achieve comparable performance. These results demonstrate that reconstructing object-level depth using our SDDR scheme and NGPH is a better choice than reconstructing the object mesh. The main reason may be it is much easier for the network to learn useful information, such as the object's shape detail or the absolute object center, if depth translation and shape points are decoupled.

\begin{figure}[t]
 \centering
  \includegraphics[width=1.0\linewidth]{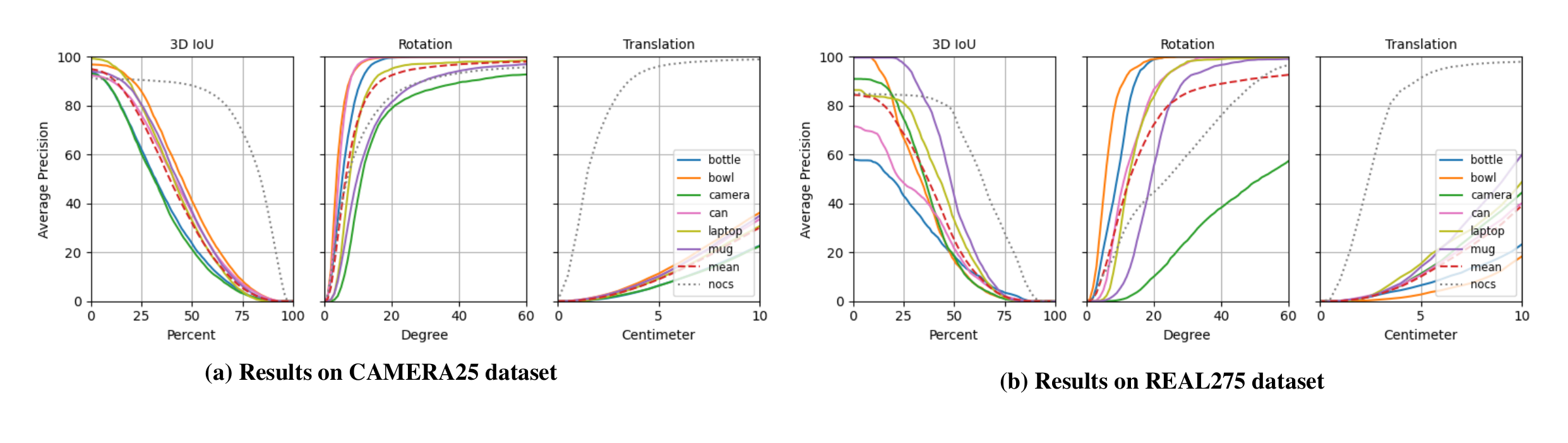}

  \caption{The average precision (AP) vs. different thresholds on 3D IoU, rotation error, and translation error.}
  
  \label{map_curve}
\end{figure}

\begin{figure*}[t]
 \centering
  \includegraphics[width=0.9\linewidth]{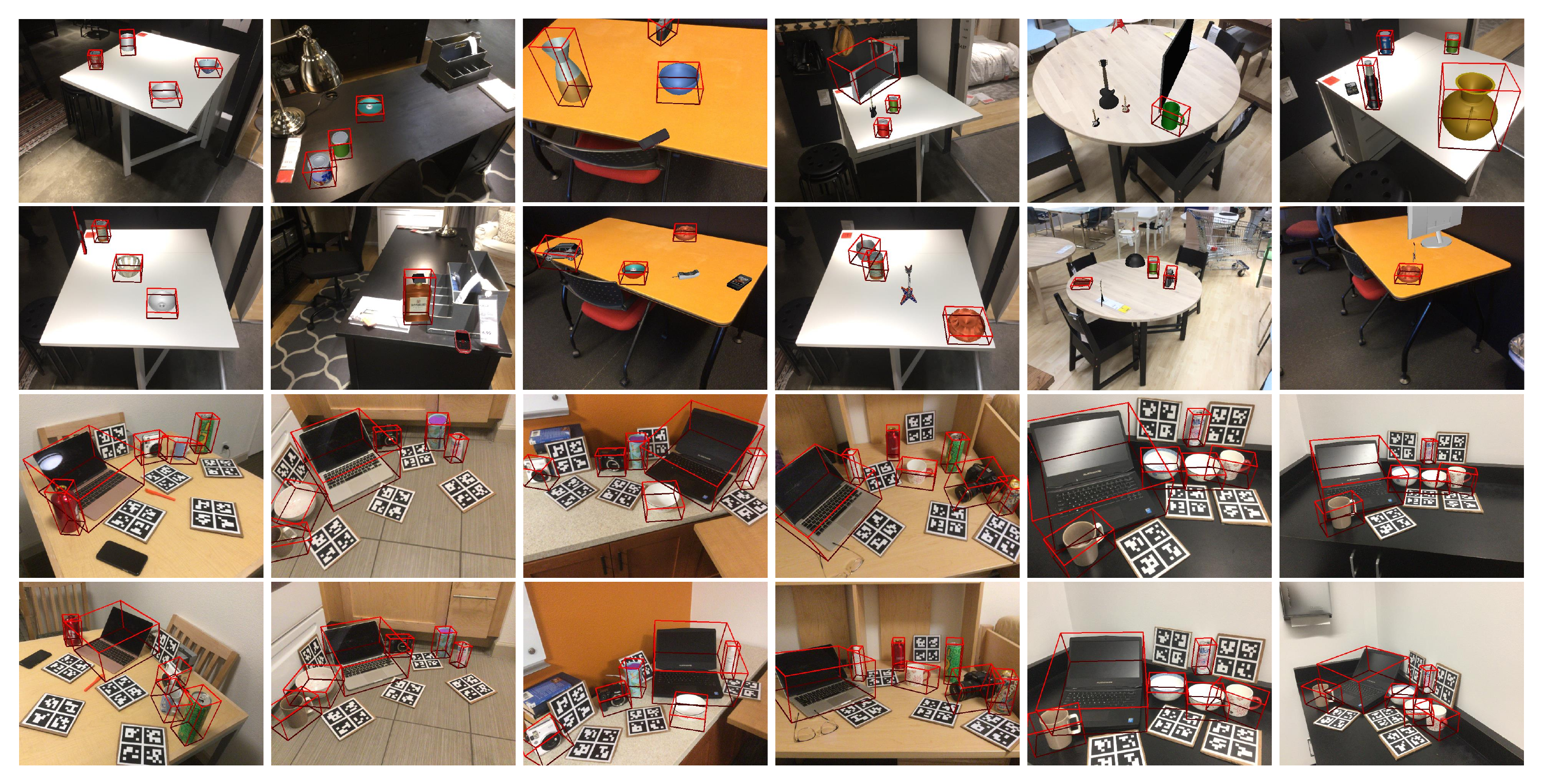}

  \caption{Qualitative results of successful cases. Top two rows are results on CAMERA25 dataset and bottom two rows are results on REAL275 dataset.	 Our method can accurately estimate the pose and size of the object taking a single RGB image as input. }
  \label{visulize_success}

\end{figure*}

\begin{table*}[t]
		\centering
		\begin{tabular}{c|c|c|c|c|c|c } 
			\hline 	
			Versions & IoU25& IoU50&IoU50& 10 cm& 10\degree & 10\degree 10cm\\	
			
			\hline 	
Vanilla	 SPD &54.9	&12.6&0.6	&6.4&55.0&4.4			\\
w/o Depth translation design		&72.7	&31.1&	\textbf{5.5}&	29.8	&74.6&	23.2		\\	
w/o Shape points design &70.0&	28.4	&4.5&	27.7	&70.8&	20.7\\	
w/o NGPH&71.2&	29.4	&4.6	&28.3&	61.6	&18.7	\\	
w/o Adversarial training	&	71.8&	29.6	&5.0&	28.3	&\textbf{75.1}&	22.1\\
Full model	 &\textbf{74.3}&	\textbf{32.1}	&5.4&\textbf{30.1}	&74.0&\textbf{23.4}\\
			\hline 	
	\end{tabular}

	\caption{Results of ablation study. We remove a module from a network architecture each time to investigate its impact.}
	\label{ablation:remove}

\end{table*}

\begin{figure}[t]
 \centering
  \includegraphics[width=0.99\linewidth]{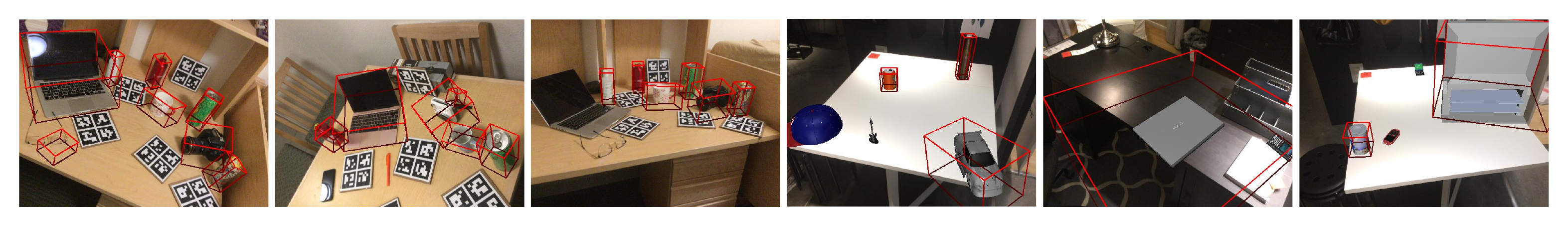}

  \caption{Qualitative results of failure cases.}
  \label{visulize_fail}

\end{figure}

Then, we compare our method with Synthesis \cite{chen2020category} and Lee et al. \cite{lee2021category} on REAL275. To avoid overfitting, we finetune our model trained on CAMERA25 train set on the REAL275 train set for 1500 steps rather than training on REAL275 from scratch. It can be seen from the table that our method is only slightly outperformed by Lee et al. \cite{lee2021category} at the IoU75 metric and the 10 cm metric. While for another 4 metrics, our model performs the best. Especially at the IoU25 metric and the 10 \degree metric, our model outperforms the second-best model for 6.7 points and 7.8 points respectively. Our model performs much better than previous methods in terms of the IoU25 metric and the 10 \degree metric mainly because the SDDR and the NGPH could equip our model with the ability of concentrating on the object's shape detail and absolute object center independently.

To further verify our motivation and the benefit of reconstructing object-level depth over estimating scene-level depth, we compare our methods with two scene-level depth estimation baselines in Table \ref{comapre:scene}. Both scene-level baseline (shared) and scene-level baseline (independent) share the same encoder-decoder network architecture. The difference is that during training, scene-level baseline (shared) shared the encoder with the NOCS reconstruction branch while scene-level baseline (independent) independently train the depth estimator. Both networks are carefully tuned to achieve their best performance. Table \ref{comapre:scene} shows that OLD-Net significantly outperforms both scene-level baselines. The reason could be that the object-level depth predicted  by OLD-Net is much better at preserving shape details than the coarse scene-level depth, while shape details are critical to the NOCS-depth aligning process. 


All the above results have demonstrated the superiority of our method compared to SOTA RGB-based methods. Moreover, we also exhibit the picture of the average precision (AP) vs. different thresholds on 3D IoU, rotation error, and translation error in Fig. \ref{map_curve}. We compare our method with NOCS \cite{wang2019normalized}, an RGBD-based method. It can be seen from the figure that our method performs excellently in terms of IoU and rotation on all categories. The excellent performance on rotation prediction largely due to that we decouple the shape points out of the depth to preserve shape detail. In alignment, whether the rotation is accurate or not is mostly dependent on the quality of the object's shape. Therefore, our model even achieves comparable performance with the RGBD-based method in terms of rotation prediction. In contrast, the translation prediction results are relatively low compared to RGBD-based methods. This is because recovering the absolute global position $z_t$ of the object from a monocular RGB image is an ill-posed problem.  The relatively inaccurate translation is the main reason why the most strict 10 \degree 10 cm metric is not high. Therefore, future works should pay more attention to obtaining more precise absolute depth prediction.

We also test the running time complexity of our method. The deep  network runs 64 FPS on a 3090 GPU, and the Umeyama algorithm runs 30FPS on a normal AMD EPYC 7702 CPU. The overall pipeline runs 22FPS.  Our method is potential of being real time. 

\noindent \textbf{Qualitative results of 6D pose estimation}: To qualitatively analyze the effectiveness of our method, we visualize the estimated bounding boxes in Fig. \ref{visulize_success}. Results on both synthetic data and real data are shown. It can be seen that OLD-Net can predict tight object bounding boxes, which are accurate enough for augmented reality products.  In Fig \ref{visulize_fail}, we also show some failure cases. OLD-Net may miss objects or detect ghosts sometimes. We leave solving it as our future work. More visualizations are in the SuppMat. 


\subsection{Ablation study}

In this section, we study the impact of our key designs by removing them out of our network architecture each time unless otherwise stated. 

\noindent \textbf{Vanilla SPD}: We adopt SPD to learn shape points in SDDR. One may wonder whether the good performance of our model comes from the SPD rather than other modules we design. Therefore, we show the performance when we only use a vanilla SPD module (without OLD-Net, without SDDR, and only directly predict the back-projected point cloud of the object-level depth using SPD). Without our other designs, the performance of the vanilla SPD is poor. This demonstrates that the SPD itself can't undertake the object-level depth prediction task. In contrast, it is the proposed SDDR and NGPH being the key components that make our method work.

\noindent \textbf{Impact of SDDR scheme}: In this paper, SDDR is introduced to decouple the object-level depth into depth translation and shape points. Compared to the Vanilla SPD, all versions of our models in Table \ref{ablation:remove} adopt the SDDR scheme, therefore, their performances are largely improved. Then, in Table \ref{ablation:remove} row 3, instead of using two separate modules to learn the depth translation and the shape points independently, we directly predict the absolute object-level depth using a single module. We find that the IoU25 metric and IoU50 metric are decreased a lot. That may be because without decoupling depth translation out, the network may lose object details like the length-width-height ratio of the object or some particular object components.   Besides, in Table \ref{ablation:remove} row 4, we show the result of replacing the SPD with a MLP to predict shape points, i.e. directly regressing NOCS and object-level depth. It is obvious all metrics are decreased significantly. The result proves that it is very necessary to adopt SPD in SDDR. The SPD provides the model with strong constraints about object shape. Note in Table \ref{ablation:remove} row 3 and row 4, though we remove some designs, the 2D coordinates of pixels belonging to the object are still utilized for back-projection (it is also a part of SDDR), which would provide additional constraints for absolute depth. Otherwise, the performance would be even worse as shown in Table \ref{ablation:remove} row 2, which directly predicts back-projected object point clouds. In summary,  the SDDR scheme plays an significant role in OLD-Net for both object shape detail preservation and absolute object center prediction.

\noindent \textbf{Impact of NGPH}: Since our model only takes an RGB image patch to predict depth to preserve shape details, the global position information would be lost. To make up for this defect, we inject the NGPH into our network. In Table \ref{ablation:remove} row 5, we remove the NGPH out of our network to investigate its impact. When it is removed, all metrics are decreased a lot. That is because, without NGPH, it is hard for the network to predict absolute depth. Though relative position between 3D points may be inferred from image patches, wrong absolute depth would make it hard to accurately recover the object pose through aligning.

\noindent \textbf{Impact of adversarial training}: We adopt an adversarial training strategy to increase the quality of the predicted NOCS representation. When it is removed, as shown in the second last row of Table \ref{ablation:remove}, all metrics except the 10 \degree metric are decreased. This result evidences that adversarial training is necessary for performance improvement. It also indicates that both the quality of the NOCS representation and the object-level depth are important. Neither can be ignored.

\section{Conclusion}
In this paper, we propose a novel network named OLD-Net for RGB-based category-level 6D object pose estimation. Directly predicting object-level depth using shape prior is the key insight of our work. To reconstruct high-quality object-level depth, we introduce the Normalized Global Position Hints and Shape-aware Decoupled Depth Reconstruction Scheme in OLD-Net. We also predict the canonical NOCS representation of the object in our pipeline using adversarial training. Extensive experiments on both real and synthetic datasets have demonstrated that our method can achieve new state-of-the-art performance. Additional limitations and future works are presented in the SuppMat.

\section{Acknowledgements}
This work was supported in part by National Key Research and Development Program of China under Grant No. 2020YFB2104101 and National Natural Science Foundation of China (NSFC) under Grant Nos. 62172421, 71771131, and 62072459.



\clearpage
%
%
\bibliographystyle{splncs04}
\bibliography{egbib}
\end{sloppypar}
\section{Appendix}
We implement our method by PyTorch and optimize it using the Adam optimizer. The image patch is resized to $192 \times 192$. We randomly select 1024 pixels to predict depth during training. The Detect-Net is Mask-RCNN \cite{he2017mask}. A PSP-Net \cite{zhao2017pyramid} with a ResNet-18 \cite{he2016deep} backbone is used to learn image features. Points number of the shape prior is 1024. We set $C=64$ and $C_g=1024$. The model is trained for 50 epochs with a batch size of 96. The initial learning rate of the main network is 0.0001 with a decay rate of 0.1 at the 40th epoch. The initial learning rate of the discriminator is 0.00001. It also decays by 0.1 at the 40th epoch. The balance terms $\gamma_1$ to $\gamma_7$ are 1.0, 0.1, 0.1, 1.0, 5.0, 0.0001 and 0.01. We train our model on a single RTX 3090 GPU. Note we also have to recover the object size, we simply use the average size of $P_{pri}+D_{nocs}$ as our result following \cite{tian2020shape}. Following \cite{lee2021category}, we report the mean Average Precision (mAP) metric. Six kinds of mAPs are chosen. They are mAP at $IoU>0.25$ (IoU25), mAP at $IoU>0.5$ (IoU50), mAP at $IoU>0.75$ (IoU75), mAP at $translation<10 cm$ (10cm), mAP at $rotation<10$\degree (10 \degree ) and mAP at the threshold of 10 \degree 10 cm.

\subsection{Datasets}
To verify the effectiveness of our method. We conduct experiments on the CAMERA25 dataset \cite{wang2019normalized} and REAL275 dataset \cite{wang2019normalized}. They are currently the most prevalent benchmark datasets for category-level 6D object pose estimation. The CAMERA25 dataset is a synthetic dataset that contains 300K RGBD images (with 25K for evaluation) generated by rendering and compositing synthetic objects into real scenes. The REAL275 dataset is a real-world dataset that contains 4.3K real-world RGBD images from 7 scenes for training, and 2.75K real-world RGBD images from 6 scenes for evaluation.  Both datasets consist of six categories, i.e., bottle, bowl, camera, can, laptop and mug. Note though they provide RGBD images, we only use the RGB part of these images to predict the 6D object pose during evaluation.

\subsection{More results}
\begin{figure*}[t]
  \centering
   \includegraphics[width=0.95\linewidth]{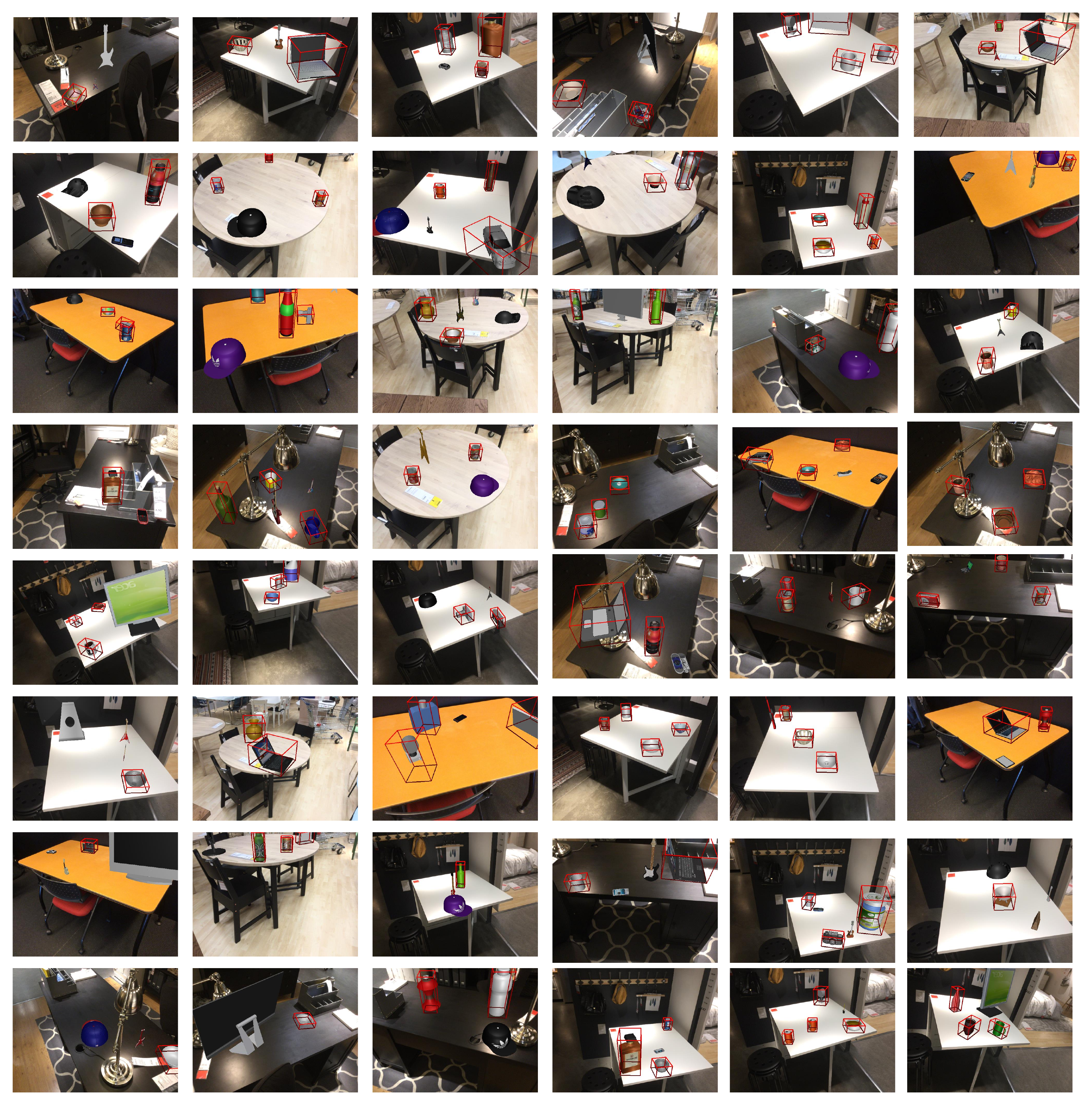}
   \caption{More visualization results on CAMERA25 dataset.}
   \label{supp_camera}
\end{figure*}

\begin{figure*}[t]
  \centering
   \includegraphics[width=0.95\linewidth]{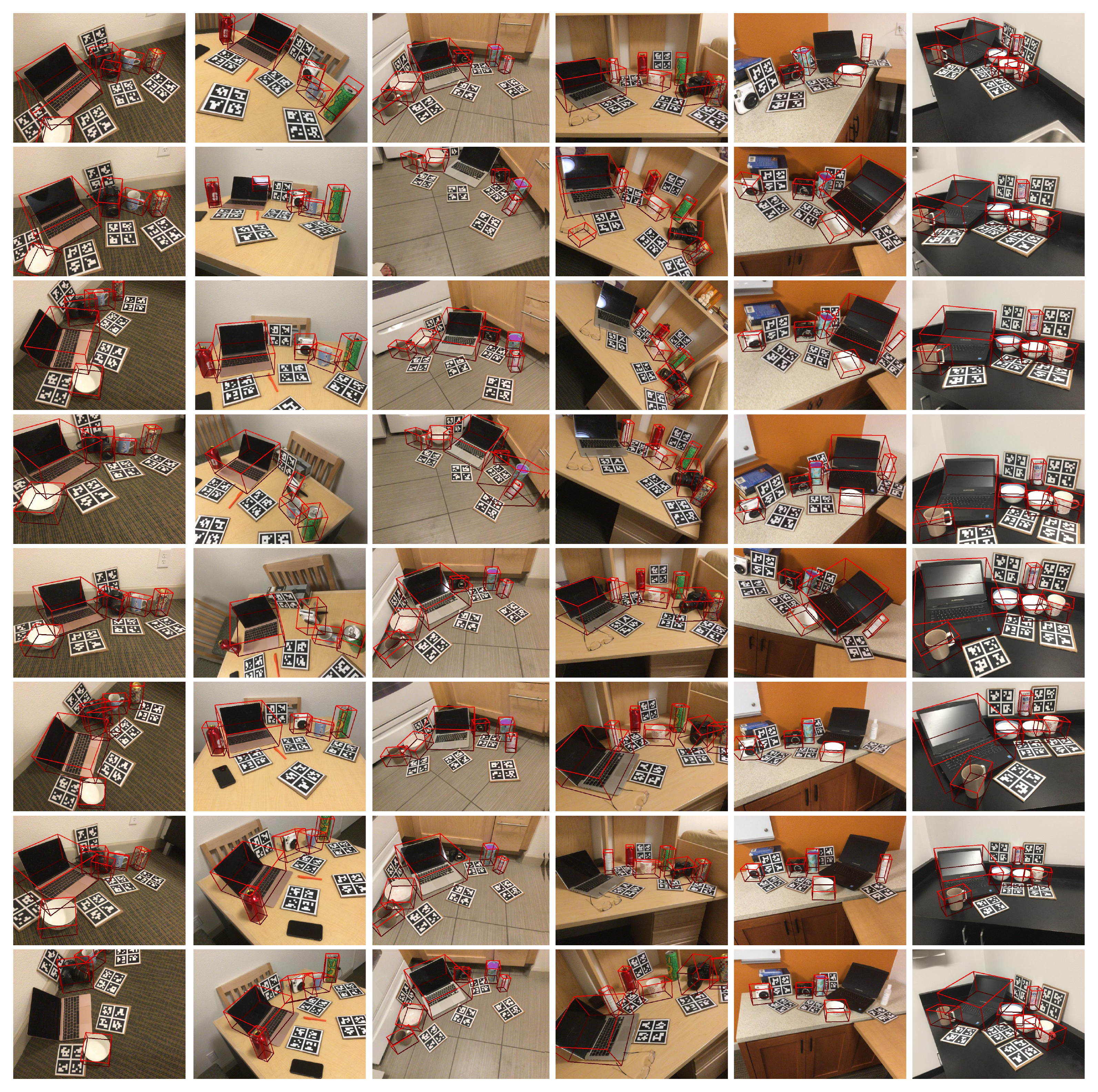}
   \caption{More visualization results on REAL275 dataset.}
   \label{supp_real}
\end{figure*}

To help readers to better understand our work, we visualize more results in Fig. \ref{supp_camera} and Fig. \ref{supp_real}.  Fig.\ref{supp_camera} shows some results on the CAMERA25 dataset and Fig. \ref{supp_real} shows some results on the REAL275 dataset. Both successful cases and failure cases are included.  We hope readers can refer to these visualizations to find more  observations that may benefit future works. 

From the failure cases we can find that that OLD-Net may miss objects or detect ghosts sometimes. This may be solved by using a stronger Detect-Net. Researching 2D detection is out of the scope of this paper and we leave it as a future work. Another observation from failure cases is that we find sometimes our model can not excellently recover the size of the object. Therefore, further future efforts should also be made to improve the object size estimation accuracy.

We believe that our work is a significant step to enable RGB-based category-level 6D object pose estimation to  be deployed into many potential applications like robotics and augmented reality.  Besides solving the limitations, there are also some future works we suggest to do: 1) Semi-supervised training on both labelled synthetic data and unlabelled real world data. 2) Designing stronger object-level depth prediction network architectures. 3) Trying domain adaptation methods.

\end{document}